\pdfoutput=1

\documentclass[11pt]{article}

\usepackage{acl}

\usepackage{times}
\usepackage{latexsym}

\usepackage[T1]{fontenc}

\usepackage[utf8]{inputenc}

\usepackage{microtype}

\title{Instructions for *ACL Proceedings}

\usepackage{times}
\usepackage{latexsym}
\usepackage{soul}
\usepackage{amsmath}
\usepackage{booktabs}
\usepackage{bm}
\usepackage{cleveref}
\usepackage{graphicx}
\usepackage{tabularx}
\usepackage{soul}
\usepackage{xcolor}
\usepackage{todonotes}
\usepackage{multirow}
\usepackage{xpatch}
\usepackage{blindtext}
\usepackage{fdsymbol}

\usepackage[T1]{fontenc}

\usepackage[utf8]{inputenc}

\usepackage{microtype}
\usepackage{hyperref}
\usepackage{adjustbox}
\usepackage{textcomp}
\usepackage{xparse}

\usepackage{color}
\usepackage{dirtytalk}
\usepackage{multirow}

\def\eg{\textit{e.g.}}

\newcommand{\myparagraph}[1]{\vspace{2pt}\noindent{\bf{#1}}~~}

\title{End-to-end Spoken Conversational Question Answering: \\ Task, Dataset and Model}

\author{Chenyu You$^{\clubsuit*}$, Nuo Chen$^{\spadesuit*}$, Fenglin Liu$^{\spadesuit}$, \\ \bf{Shen Ge$^{\heartsuit}$,} Xian Wu$^{\heartsuit}$, Yuexian Zou$^{\spadesuit}$ \\ \textit{}
$^\clubsuit$Yale University $^\heartsuit$Peking University $^\spadesuit$Tencent \\
  \texttt{chenyu.you@yale.edu}~~~ \texttt{\{nuochen,fenglinliu98,zouyx\}@pku.edu.cn} \\ \texttt{\{shenge,kevinxwu\}@tencent.com} \\
}

\begin{document}
\maketitle
\renewcommand{\thefootnote}{\fnsymbol{footnote}}
\footnotetext[1]{Equal contribution.}
\renewcommand{\thefootnote}{\arabic{footnote}}

\begin{abstract}
In spoken question answering, the systems are designed to answer questions from contiguous text spans within the related speech transcripts. However, the most natural way that human seek or test their knowledge is via human conversations. Therefore, we propose a new \textbf{S}poken \textbf{C}onversational \textbf{Q}uestion \textbf{A}nswering task (SCQA), aiming at enabling the systems to model complex dialogue flows given the speech documents. In this task, our main objective is to build the system to deal with conversational questions based on the audio recordings, and to explore the plausibility of providing more cues from different modalities with systems in information gathering. To this end, instead of directly adopting automatically generated speech transcripts with highly noisy data, we propose a novel unified data distillation approach,~\textsc{DDNet}, which effectively ingests cross-modal information to achieve fine-grained representations of the speech and language modalities. Moreover, we propose a simple and novel mechanism, termed Dual Attention, by encouraging better alignments between audio and text to ease the process of knowledge transfer. To evaluate the capacity of SCQA systems in a dialogue-style interaction, we assemble a \textbf{Spoken} \textbf{Co}nversational \textbf{Q}uestion \textbf{A}nswering (Spoken-CoQA) dataset with more than 40k question-answer pairs from 4k conversations. The performance of the existing state-of-the-art methods significantly degrade on our dataset, hence demonstrating the necessity of cross-modal information integration. Our experimental results demonstrate that our proposed method achieves superior performance in spoken conversational question answering tasks.
\end{abstract}

\begin{figure*}
\begin{center}
\includegraphics[width=\textwidth]{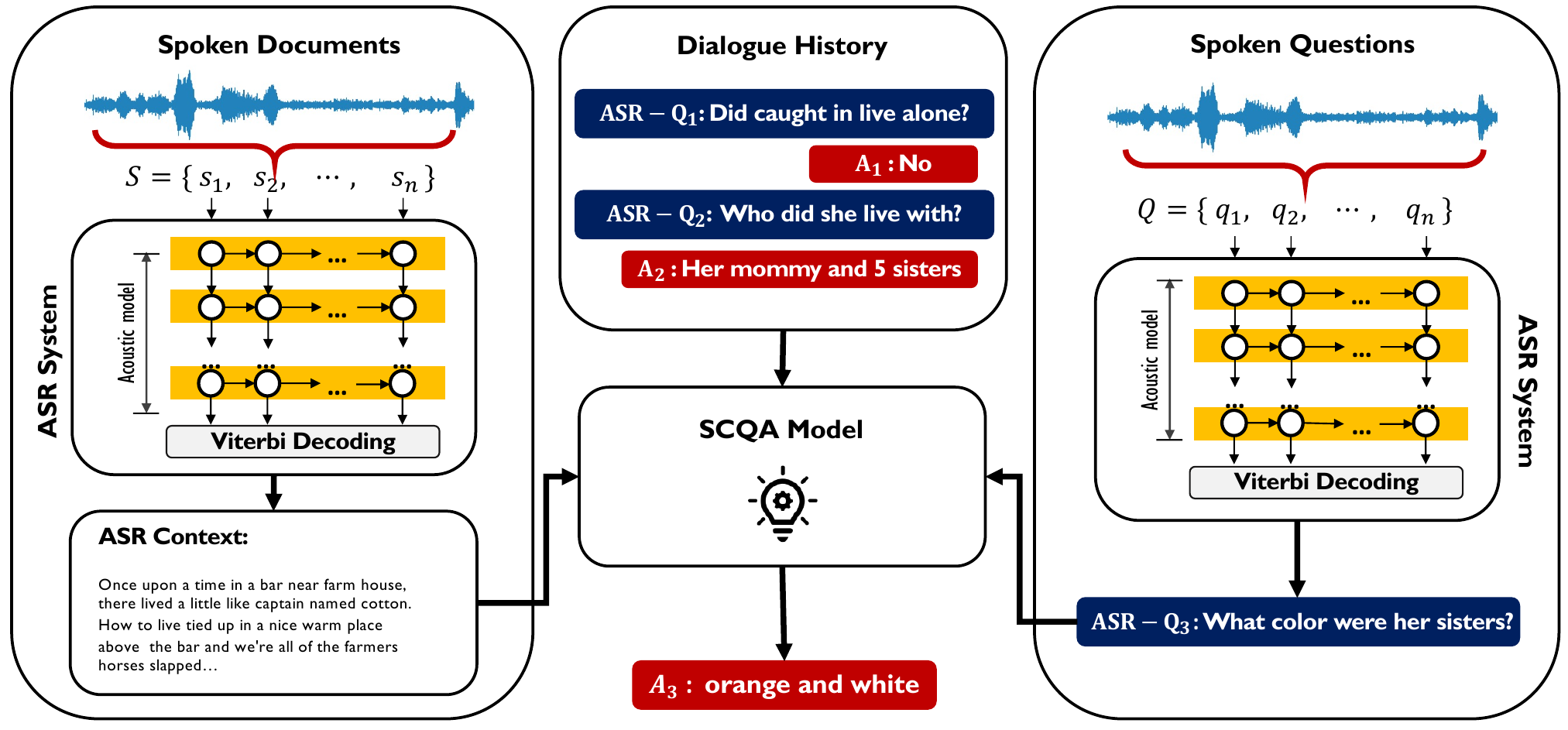} 
\end{center}
\vspace{-10pt}
\caption{An illustration of flow diagram for spoken conversational question answering tasks with an example from our proposed Spoken-CoQA dataset.}
\label{fig:framework}
\vspace{-5pt}
\end{figure*}

\begin{table*}[t]
\begin{center}

\vspace{-5pt}
\scriptsize

\begin{tabular}{ll}
\toprule
\multicolumn{1}{c}{\bf Manual Transcript}  &\multicolumn{1}{c}{\bf ASR Transcript}
\\ \midrule
\begin{minipage}[t]{0.45\textwidth}%
Once upon a time, in a barn near a farm house, there lived a little white kitten named Cotton. Cotton lived high up in a nice warm place above the barn where all of the farmer's horses slept. But Cotton wasn't alone in her little home above the barn, oh no. She shared her hay bed with her mommy and 5 other sisters\dots
\end{minipage}
&
\begin{minipage}[t]{0.45\textwidth}%
Once upon a time in a \textbf{bar} near farm house, there lived a little~\textbf{like captain} named {cotton}.~\textbf{How to live} tied up in a nice warm place above  the~\textbf{bar} and~\textbf{we're} all of the farmers horses \textbf{slapped}. But~\textbf{caught in} was not alone in her little home above the bar~\textbf{in now}. She shared her hey bed with her mommy and 5 other sisters\dots
\end{minipage}
\\ \\
\begin{minipage}[t]{0.45\textwidth}%
Q$_{1}$: Did Cotton live alone? \\
A$_{1}$: no  \\
R$_{1}$: Cotton wasn't alone.
\end{minipage}
&
\begin{minipage}[t]{0.45\textwidth}%
ASR-Q$_{1}$: Did \textbf{caught in} live alone? \\
A$_{1}$: no  \\
R$_{1}$: \textbf{Caught} \textbf{in} wasn't alone.
\end{minipage}
\\ \\
\begin{minipage}[t]{0.45\textwidth}%
Q$_{2}$: Who did she live with? \\
A$_{2}$: with her mommy and 5 sisters  \\
R$_{2}$: with her mommy and 5 other sisters
\end{minipage}
&
\begin{minipage}[t]{0.45\textwidth}%
ASR-Q$_{2}$: Who did she live with?    \\
A$_{2}$: with her mommy and 5 sisters  \\
R$_{2}$: with her mommy and 5 other sisters
\end{minipage}
\\ \\
\begin{minipage}[t]{0.45\textwidth}%
Q$_{3}$: What color were her sisters? \\
A$_{3}$: orange and white  \\
R$_{3}$: her sisters were all orange with beautiful white tiger stripes
\end{minipage}
&
\begin{minipage}[t]{0.45\textwidth}%
ASR-Q$_{3}$: What color were her sisters?    \\
A$_{3}$: orange and white  \\
R$_{3}$: her sisters were all orange with beautiful white tiger stripes
\end{minipage}
\\ \bottomrule
\end{tabular}
\end{center}
\caption{An example from Spoken-CoQA. We can observe large misalignment between the manual transcripts and the corresponding ASR transcripts. Note that the misalignment is in~\textbf{bold} font and the example is the extreme case. For more dataset information, please see Section \ref{subsec:data} and Appendix Section ``More Information about Spoken-CoQA''.}
\label{tab:scqa_example}
\vspace{-10pt}
\end{table*}

\begin{figure*}[t]
\centering
\includegraphics[width=\linewidth]{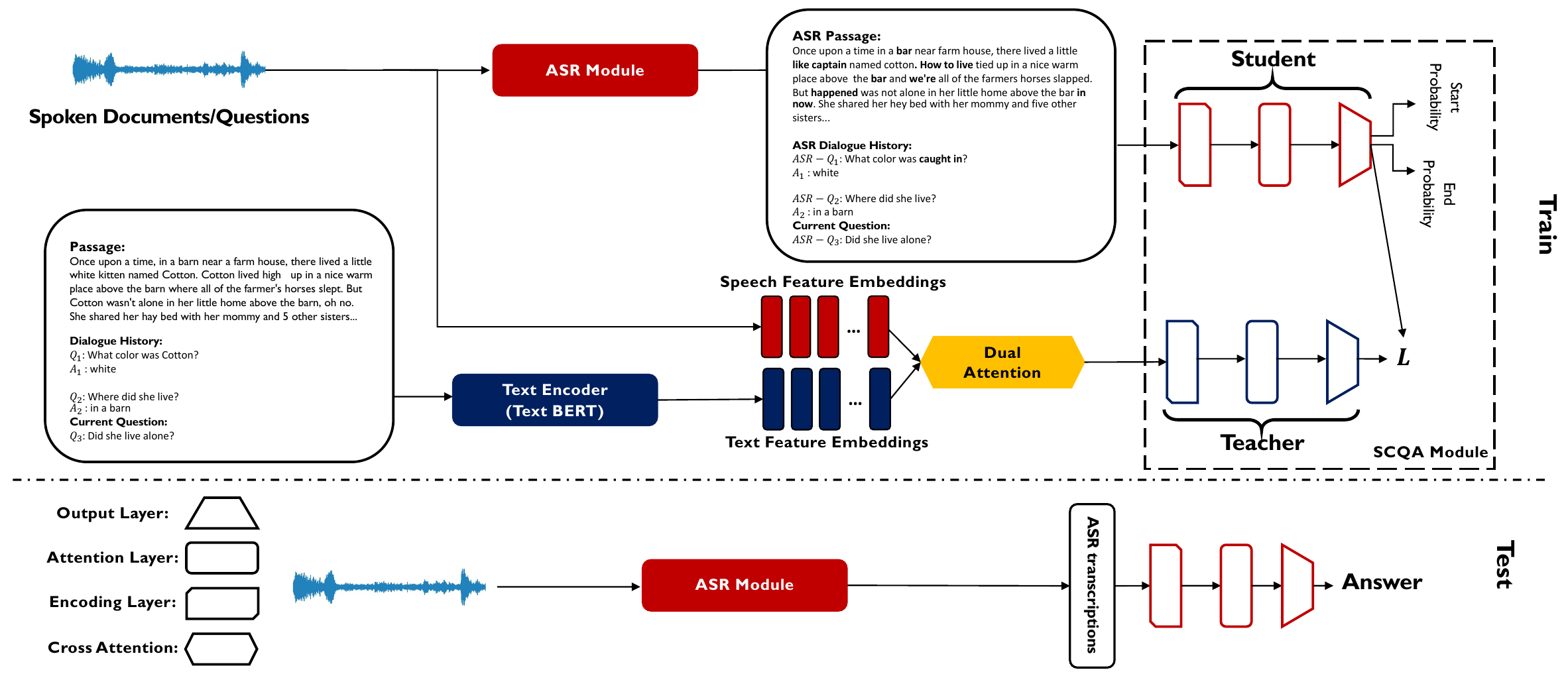}
\vspace{-10pt}
\caption{An illustration of the architecture of \textsc{DDNet}. In training stage, we adopt the teacher-student paradigm to enable the student model (only trained on speech documents) to achieve good performance. As for test, we only use student model for inference.}
\label{fig:model}
\vspace{-15pt}
\end{figure*}

\section{Introduction}

Conversational question answering (CQA) has been studied extensively over the past few years within the natural language processing (NLP) communities~\cite{zhu2018sdnet,liu2019roberta,yang2019xlnet}. Different from traditional question answering (QA) tasks, CQA aims to enable models to learn the representation of the context paragraph and multi-turn dialogues. Existing CQA methods \cite{huang2018flowqa,devlin2018bert,xu2019review,gong2020recurrent} have achieved superior performances on several benchmark datasets, such as  QuAC~\cite{choi2018quac} and CoQA~\cite{elgohary2018dataset}.

Current CQA research mainly focuses on leveraging written text sources in which the answer can be extracted from a large document collection. However, humans communicate with each other via spontaneous speech (\eg, meetings, lectures, online conversations), which convey rich information. Consider our multimodal experience, fine-grained representations of both audio recordings and text documents are considered to be of paramount importance. Thus, we learn to draw useful relations between modalities (speech and language), which enables us to form fine-grained multimodal representations for end-to-end speech-and-language learning problems in many real-world applications, such as voice assistant and chat robot.

In this paper, we propose a novel and challenging spoken conversational question answering task - SCQA. An overview pipeline of this task is shown in Figure~\ref{fig:framework}. Collecting such a SCQA dataset is a non-trivial task, as in contrast to current CQA tasks, we build our SCQA with two main goals as follows: (1) SCQA is a multi-turn conversational spoken question answering task, which is more challenging than \textit{only} text-based task; (2) existing CQA methods rely on a single modality (text) as the context source. However, plainly leveraging uni-modality information is naturally undesirable for end-to-end speech-and-language learning problems since the useful connections between speech and text are elusive. Thus, employing data from the context of another modality (speech) can allow us to form fine-grained multimodal representations for the downstream speech-and-language tasks; and (3) considering the speech features are based on regions and are not corresponding to the actual words, this indicates that the semantic inconsistencies between the two domains can be considered as the \textit{semantic gap}, which requires to be resolved by the downstream systems themselves.

In order to provide a strong baseline for this challenging multi-modal spoken conversational question answering task, we first present a novel knowledge distillation (KD) method for the proposed SCQA task. Our intuition is that speech utterances and text contents share the dual nature property, and we can take advantage of this property to learn the correspondences between these two forms. Specifically, we enroll multi-modal knowledge into the~\textit{teacher} model, and then guide the~\textit{student} (\textit{only} trained on noisy speech documents) to boost network performance. Moreover, considering that the semantics of the speech features and the textual representations are usually inconsistent, we introduce a novel mechanism, termed Dual Attention, to encourage fine-grained alignments between audio and text to close the cross-modal semantic gap between speech and language. One example of cross-modal gap is shown in Table~\ref{tab:scqa_example}. The experimental results show that our proposed \textsc{DDNet} achieves remarkable performance gains in the SCQA task. To the best of our knowledge, we are the first work in spoken conversational question answering task.

Our main contributions are as follows:
\begin{itemize}
    \item We propose Spoken Conversational Question Answering task (SCQA), and comprise Spoken-CoQA dataset for machine comprehension of spoken question-answering style conversations. To the best of our knowledge, our Spoken-CoQA is the first spoken conversational question answering dataset.\vspace{-5pt}
    \item We develop a novel end-to-end method based on data distillation to learn both from speech and language domain. Specifically, we use the model trained on clear texts as well as recordings to guide the model trained on noisy speech transcriptions. Moreover, we propose a novel Dual Attention mechanism to align the speech features and textual representations in each domain.\vspace{-5pt}
    \item We demonstrate that, by applying our proposed \textsc{DDNet} on several previous baselines, we can obtain considerable performance gains on our proposed Spoken-CoQA dataset.
\end{itemize}

\section{Related Work}
\myparagraph{Text Question Answering.}
In recent years, the natural language processing research community has devoted substantial efforts to text question answering tasks \cite{huang2018flowqa,zhu2018sdnet,xu2019review,zhang2019sgnet,gong2020recurrent,chen2020adaptive}. Within the growing body of work on machine reading comprehension, an important sub-task of text question answering, two signature attributes have emerged: the availability of large benchmark datasets \cite{choi2018quac, elgohary2018dataset,reddy2019coqa} and pre-trained language models \cite{devlin2018bert,liu2019roberta,lan2019albert}. However, these existing works typically focus on modeling the complicated context dependency in text form. In contrast, we focus on enabling the machine to build the capability of language recognition and dialogue modeling in both speech and text domains.

\myparagraph{Spoken Question Answering.}
In parallel to the recent works in natural language processing \cite{huang2018flowqa,zhu2018sdnet}, these trends have also been pronounced in the speech field \cite{chen2018spoken,haghani2018audio,lugosch2019speech,palogiannidi2020end,you2020contextualized,you2021knowledge,you2021mrd,you2021self,you2020data,chen2021self,xu2021semantic,su2020audeo,su2021does}, where spoken question answering (SQA), an extended form of QA, has explored the prospect of machine comprehension in spoken form. Previous work on SQA typically includes two separate modules: automatic speech recognition (ASR) and text question answering. It involves transferring spoken content to ASR transcriptions, and then employs NLP techniques to handle speech tasks. 

Existing methods \cite{tseng2016towards,serdyuk2018towards,su2020improving} focus on optimizing each module in a two-stage manner, where errors in the ASR module would result in severe performance loss. \citet{lee2019mitigating} proved that utilizing clean texts can help model trained on the ASR transcriptions to boost the performance via domain adaptation. \citet{chuang2019speechbert} cascaded the BERT-based models as a unified model, and then trained it in a joint manner of audio and text. However, the existing SQA methods aimed at solving a single question given the related passage, without building and maintaining the connections of different questions in the human conversations. In addition, we compare our Spoken-CoQA with existing SQA datasets (See Table~\ref{table:sqa-vs-scqa}). Unlike existing SQA datasets, Spoken-CoQA is a multi-turn conversational SQA dataset, which is more challenging than single-turn benchmarks.

\myparagraph{Knowledge Distillation.}
\citet{hinton2015distilling} introduced the idea of Knowledge Distillation~(KD)~in a~\textit{teacher-student} scenario. In other words, we can distill the knowledge from one model (massive or~\textit{teacher} model) to another (small or~\textit{student} model). Previous work has shown that KD can significantly boost prediction accuracy in natural language processing and speech processing \cite{kim2016sequence,hu2018attention,huang2018knowledge,hahn2019self,liu2021auto,liu2021aligning,cheng2016random,cheng2016hybrid,cheng2016identification,you2020unsupervised,you2021momentum,you2022simcvd,you2022class,you2018structurally,you2019ct,you2019low,lyu2018super,lyu2019super,guha2020deep,yang2020nuset,ma2021good,ma2021undistillable}, while adopting KD-based methods for SQA tasks has been less explored. In this work, our goal is to handle the SCQA tasks. More importantly, we focus the core nature property in speech and text: Can spoken conversational dialogues further assist the model to boost the performance? Finally, we incorporate the knowledge distillation framework to distill reliable dialogue flow from the spoken contexts, and utilize the learned predictions to guide the \textit{student} model to train well on the noisy input data.

\begin{table}[t]
\centering

\resizebox{0.45\textwidth}{!}{%
\begin{tabular}{lccc}
\bottomrule
\multicolumn{1}{l}{\bf Dataset}  &\multicolumn{1}{c}{\bf Conversational} &\multicolumn{1}{c}{\bf Spoken}  &\multicolumn{1}{c}{\bf Answer Type}
\\ \midrule
TOEFL \cite{tseng2016towards}  & $\times$ & $\surd$ & Multi-choice \\
S-SQuAD \cite{li2018spoken}  &  $\times$  &  $\surd$& Spans   \\
ODSQA \cite{lee2018odsqa}&  $\times$  & $\surd$ &  Spans    \\
\midrule
S-CoQA & $\surd$ & $\surd$ & Free-form \\
\bottomrule
\end{tabular}
}
\caption{Comparison of Spoken-CoQA with existing spoken question answering datasets. S-SQuAD and S-CoQA denote Spoken-SQuAD and Spoken-CoQA, respectively.}
\label{table:sqa-vs-scqa}
\vspace{-10pt}
\end{table}

\section{Task Definition}

In this section, we propose the novel SCQA task and collect a Spoken-CoQA (S-CoQA) dataset, which uses the spoken form of multi-turn dialogues and spoken documents to answer questions in multi-turn conversations.

Given a spoken document $D^s$, we use $D^t$ to denote the clean original text and $D^a$ to denote the ASR transcribed document. We also have $Q^a_{1:L}$=\{$q^{a}_{1},q^{a}_{2},...,q^{a}_{L}$\}, which is a collection of $L$-turn ASR transcribed spoken questions $Q^s_{1:L}$, as well as $A^{t}_{1:L}$= \{$a^{t}_{1},a^{t}_{2},...,a^{t}_{L}$\} which are the corresponding answers to the questions in clean texts. The objective of SCQA task is then to generate the answer $a^t_L$ for question $q^a_L$, given document $D^a$, multi-turn history questions $Q^a_{1:L-1}$=\{$q^{a}_{1},q^{a}_{2},...,q^{a}_{L-1}$\}, and reference answers $A^t_{1:L-1}$= \{$a^{t}_{1},a^{t}_{2},...,a^{t}_{L-1}$\}. In other words, our task in the testing phase can be formulated as
\begin{equation}
\begin{aligned}
\{D^s,Q^s_{1:L}\}\xrightarrow{\text{ASR}}\{q^a_L,D^a,Q^a_{1:L-1},a^t_{1:L-1}\} &\to a^t_L
\label{eqn:task_def}
\end{aligned}
\end{equation}

Please note that in order to improve the performance, in the training phase, we make use of auxiliary information which are the clean texts of document $D^t$ and dialogue questions $Q^t$=\{$q^t_1,q^t_2,\ldots,q^t_L$\}, to guide the training of student model. As a result, the training process could be formulated as below:
\begin{equation*}
\footnotesize\left.
\begin{aligned}
{\tiny\text{student:}\ } \{D^s,Q^s_{1:L}\}\xrightarrow{\text{ASR}}\{q^a_L,D^a,Q^a_{1:L-1},a^t_{1:L-1}\} &\\
\hfill{\tiny\text{teacher:}\ } \{D^t,Q^t_{1:L}\} &
\label{eqn:task_def}
\end{aligned}
\right\} \to a^t_L
\end{equation*}
However, in the inference stage, these additional information of $D^t$ and $Q^t_{1:L}$ are not needed.

\vspace{-0.3em}
\section{DDNet}
\vspace{-0.3em}

In this section, we propose \textsc{DDNet} to deal with the SCQA task, which is illustrated in Figure \ref{fig:model}. We first describe the embedding generation process for both audio and text data. Next, we propose {\em Dual Attention} to fuse the speech and textual modalities. After that, we present the major components of the \textsc{DDNet} module. Finally we describe a simple yet effective distillation strategy in the proposed \textsc{DDNet} to learn enriched representations in the speech-text domain comprehensively.

\vspace{-0.3em}
\subsection{Embedding}
\vspace{-0.3em}
\label{subsec:cmm}
Given spoken words $S$ = $\{{s_1},{s_2},...,{s_m}\}$  and corresponding clean text words $T$ = $\{{t_1},{t_2},...,{t_n}\}$, we utilize Speech-BERT and Text-BERT to generate speech feature embedding~$ \textbf{E}_s$=$\{ \textbf{E}_{s1}, \textbf{E}_{s2},..., \textbf{E}_{sm}\}$ and context word embedding~$ \textbf{E}_t$=$\{ \textbf{E}_{t1},\textbf{E}_{t2},...,\textbf{E}_{tn}\}$\footnote{In our implement, the padding strategy is used to keep m and n to be the same as the max sequence length.}, respectively. Concretely, for speech input, we first use vq-wav2vec~\cite{baevski2019vq} to transfer speech signals into a series of tokens, which is the standard tokenization procedure in speech related tasks. Next, use Speech-BERT~\cite{chuang2019speechbert}, a variant of BERT-based models retrained on our Spoken-CoQA dataset, to process the speech sequences for training.
The text contents are embbed into a sequence of vectors via our text encoder - Text-BERT, with the same architecture of BERT-base~\cite{devlin2018bert}.

\vspace{-0.3em}
\subsection{Dual Attention}
\vspace{-0.3em}
Dual Attention (DA) is proposed to optimize the alignment between speech and language domains by capturing useful information from the two domains. In particular, we first use cross attention to align speech and text representations in the initial stage. After that, we utilize contextualized attention to further align the cross-modal representations in the contextualized word-level. Finally, we employ the self-attention mechanism to form fine-grained audio-text representations.

\myparagraph{Cross Attention.}
Inspired by ViLBERT~\cite{lu2019vilbert}, we apply the co-attention transformer layer, a variant of Self-Attention~\cite{vaswani2017attention}, as the Cross Attention module for the fusing of speech and text embeddings. The Cross Attention is implemented by the standard Attention module involving Multi-Head Attention (MHA) and Feed-Forward Network (FFN) \cite{vaswani2017attention} as below:
\begin{align}
\begin{split}
\text{Attention}(Q,K,V)&=\text{FFN}(\text{MHA}(Q,K,V))\\
\text{CrossAttention}(F_1,F_2)&=\text{Attention}(F_1,F_2,F_2)
\end{split}
\end{align}
where $Q$, $K$, $V$ denote query, key, and value matrices, and $F_1$, $F_2$ denote features from difference modalities, respectively. The co-attention module then use the Cross Attention function to compute the cross attention-pooled features, by querying one modality using the query vector of another modality.
\begin{equation}
\begin{split}
\hat{ \textbf{E}}{_s^{\text{cross}}} &= \text{CrossAttention}(\textbf{E}_s,\textbf{E}_t) \\
&= \text{Attention}( \textbf{E}_s, \textbf{E}_t, \textbf{E}_t), \\
\hat{ \textbf{E}}{_t^{\text{cross}}} &= \text{CrossAttention}( \textbf{E}_t,\textbf{E}_s) \\
&=\text{Attention}( \textbf{E}_t, \textbf{E}_s, \textbf{E}_s),
\end{split}
\end{equation}
where $\hat{ \textbf{E}}{_s^{\text{cross}}} \in \mathbb{R}^{n\times d}$, $\hat{ \textbf{E}}{_t^{\text{cross}}} \in \mathbb{R}^{n\times d}$ and $d$ is the dimension of feature vectors.

\myparagraph{Contextualized Attention (CA).}
After obtaining \textit{speech-aware} representation $\hat{ \textbf{E}}{_s^\text{cross}}$ and \textit{text-aware} representation $\hat{ \textbf{E}}{_t^\text{cross}}$, our next goal is to construct more robust contextualized cross-modal representations by integrating features from both modalities. The features with fused modalities are computed as follows:
\begin{equation}
    \textbf{H}_{\text{CA}}\!=\!\text{ReLU}(\hat{ \textbf{E}}{_s^{\text{cross}}}W_1^T)\text{ReLU}(\hat{ \textbf{E}}{_x^{\text{cross}}}W_1^T)W_2^T,
\end{equation}
where $W_1,W_2 \in \mathbb{R}^{n\times d}$ are trainable weights.

\myparagraph{Self-Attention.}
To build a robust SCQA system, special attention needs to be paid on the sequential order of the dialogue, since the changes in utterances order may cause severely low-quality and in-coherent corpora. As a result, to capture the long-range dependencies such as co-references for the downstream speech-and-language tasks, similar to \cite{li2016diversity,zhu2018sdnet}, we introduce a self-attention layer to obtain the final Dual Attention (DA) representations.
\begin{equation}
\begin{split}
   \textbf{E}_{\text{DA}} &= \text{SelfAttention}(\textbf{H}_{\text{CA}}) \\
   &= \text{Attention}(\textbf{H}_{\text{CA}}, \textbf{H}_{\text{CA}},  \textbf{H}_{\text{CA}}).
\end{split}
\end{equation}

\subsection{Key Components}
\vspace{-0.3em}
\label{subsec:cmrc}
The framework of our SCQA module is similar to recent works~\cite{zhu2018sdnet,huang2017fusionnet}, which is divided into three key components: Encoding Layer, Attention Layer and Output Layer.

\myparagraph{Encoding Layer.}
Then documents and conversations (questions and answers) are first converted into the corresponding feature embeddings (i.e., character embeddings, word embeddings, and contextual embedding). The output contextual embeddings are then concatenated by the aligned cross-modal embedding~$ \textbf{E}_{\text{DA}}$ to form the encoded input features:
\begin{equation}
     \textbf{E}_{enc}=[ \textbf{E}_{t}; \textbf{E}_{\text{DA}}].
\end{equation}

\myparagraph{Attention Layer.}
We compute the attention on the context representations of the documents and questions, and extensively exploit correlations between them. Note that we adopt the default attention layers in four baseline models.

\myparagraph{Output Layer.}
After obtaining attention-pooled representations, the Output Layer computes the probability distributions of the start and end index within the entire documents and predicts an answer to current question:
\begin{equation}
    \mathcal{L}=
     -\texttt{log}\mathcal{P}(\text{st}=a_{L,\text{st}}|\mathbf{X}) - \texttt{log}\mathcal{P}(\text{ed}=a_{L,\text{ed}}|\mathbf{X})
\end{equation}
where $\mathbf{X}$ denotes the input document $D$ and $Q^{L}$, and ``$\text{st}$'', ``$\text{ed}$'' denote the start and end positions.

\vspace{-0.5em}
\subsection{Knowledge Distillation}
\vspace{-0.5em}
\label{subsec:kd}
In previous speech-language models, the only guidance is the standard training objective to measure the difference between the prediction and the reference answer. However, for noisy ASR transcriptions, such criteria may not be suitable enough. To overcome such problem, we distill the knowledge from our~\textit{teacher} model, and use them to guide the~\textit{student} model to learn contextual features in our SCQA task. Concretely, we set the model trained on the speech document and the clean text corpus as the~\textit{teacher} model and trained on the ASR transcripts as the~\textit{student} model, respectively. Thus, the~\textit{student} trained on low-quality data learns to absorb the knowledge that the~\textit{teacher} has discovered. Given the $z_{S}$ and $z_{T}$ as the prediction vectors by the~\textit{student} and~\textit{teacher} models, the objective is defined as:
\begin{equation}
    \mathcal{L}_{\text{SCQA}} \!\!=\!\!\sum_{x\in \mathcal{X}} (\alpha \tau^2 \text{KL}(p_{\tau}(z_{S}), p_{\tau}(z_{T})) + (1 - \alpha) \mathcal{L}),
\end{equation}
where $\text{KL}(\cdot)$ denotes the Kullback-Leibler divergence. $p_{\tau}(\cdot)$ is the softmax function with temperature $\tau$, and $\alpha$ is a  balancing factor.

\begin{table}[t]
   
    \vspace{-10pt}
    \centering
    \resizebox{0.45\textwidth}{!}{%
    \begin{tabular}{c c c c c}
    \toprule
        Domain & Passages & QA-Pairs& Passage Length& Avg.Turns\\
        \hline
       Children & 357 & 3.5k & 212&9.8\\
       Literature &898&8.7k&275&9.7\\
        Mid./High School&878&9.5k&308&10.8\\
         News&967&9.5k&271&9.8\\
       Wikipedia&864&8.9k&249&10.3\\
\hline
Overall&3964&40.1k&270&10.1\\
\bottomrule
    \end{tabular}
    }
      \caption{Statistical analysis on Spoken-CoQA.}
    \label{tab:stats}
    \vspace{-10pt}
\end{table}

\begin{table*}[t]
\small
\centering

  \vspace{-10pt}
    \resizebox{0.6\textwidth}{!}{%
    \begin{tabular}{l | c c|c c|c c|c c}
        \toprule
        &\multicolumn{4}{c}{\textbf{CoQA}}&
        \multicolumn{4}{c}{\textbf{S-CoQA}}\\
        \toprule\toprule
        &\multicolumn{2}{c}{\textbf{CoQA dev}}&
        \multicolumn{2}{|c}{\textbf{S-CoQA test}}&\multicolumn{2}{|c|}{\textbf{CoQA dev}}&
        \multicolumn{2}{|c}{\textbf{S-CoQA test}}\\
        \textbf{Methods} &EM &F1 &EM &F1 & EM&F1 &EM&F1 \\
        \midrule\midrule
        FlowQA \cite{huang2018flowqa} & 66.8& 75.1 & 44.1& 56.8 & 40.9& 51.6 & 22.1& 34.7  \\
        SDNet \cite{zhu2018sdnet} & 68.1 & 76.9  & 39.5 & 51.2 & 40.1 & 52.5  & 41.5 & 53.1\\
        BERT-base~\cite{devlin2018bert} & 67.7 & 77.7  & 41.8 & 54.7 & 42.3 & 55.8  & 40.6 & 54.1\\
        ALBERT-base~\cite{lan2019albert}  & 71.4&80.6 &42.6& 54.8 & 42.7&56.0 &41.4& 55.2\\
        \midrule
        Average  & 68.5 &77.6 &42& 54.4 & 41.5 & 54.0 & 36.4 & 49.3\\
        \bottomrule
  \end{tabular}
  }
    \caption{Comparison of four baselines~(FlowQA, SDNet, BERT, ALBERT). Note that we denote Spoken-CoQA test set as S-CoQA test for brevity.}
    \label{tab:my_label_2}
    \vspace{-5pt}
\end{table*}

\begin{table*}[ht]
    \small
    \centering
   
    \vspace{-15pt}
    \resizebox{0.85\textwidth}{!}{%
    \begin{tabular}{l|ccc|ccc}
        \toprule
    & \multicolumn{3}{|c|}{\textbf{CoQA dev}}&\multicolumn{3}{|c}{\textbf{S-CoQA test}}\\
    \textbf{Methods}&EM&F1&AOS &EM&F1&AOS\\
    \toprule\toprule
    FlowQA \cite{huang2018flowqa}&40.9&51.6&30.6&22.1&34.7&16.7\\
    FlowQA + sub-word unit \cite{li2018spoken}&41.9&53.2&31.4&23.3&36.4&17.4\\
    FlowQA+ SLU \cite{serdyuk2018towards}&41.2&52.0&30.6&22.4&35.0&17.1\\
    FlowQA + back-translation \cite{lee2018odsqa}& 40.5&52.1&30.8&22.9&35.8&17.3\\
    FlowQA + domain adaptation \cite{lee2019mitigating}& 41.7&53.0&31.8&23.4&36.1&17.7\\
    FlowQA + \textbf{Dual Attention}&42.3&53.0&32.7&23.5&38.8&18.9\\
    FlowQA + \textbf{Knowledge Distillation} &42.5&53.7&32.1&23.9&39.2&18.4\\
    FlowQA + \textbf{Dual Attention}+\textbf{Knowledge Distillation} & \textbf{44.3}&\textbf{55.9}&\textbf{34.4}&\textbf{26.3}&\textbf{42.4}&\textbf{21.1}\\
    \midrule
    SDNet \cite{zhu2018sdnet}&40.1&52.5&41.1&41.5&53.1&42.6\\
    SDNet + sub-word unit \cite{li2018spoken}&41.2&53.7&41.9&41.9&54.7&43.4\\
    SDNet+ SLU \cite{serdyuk2018towards}&40.2&52.9&41.2&41.7&53.2&42.6\\
    SDNet + back-translation \cite{lee2018odsqa}& 40.5&53.1&41.5&42.4&54.0&42.9\\
    SDNet + domain adaptation \cite{lee2019mitigating}& 41.0&53.9&42.0&41.7&54.6&43.6\\
    SDNet + \textbf{Dual Attention}&41.7&55.2&43.4&43.2&56.1&44.2\\
    SDNet + \textbf{Knowledge Distillation}&41.7&55.6&43.6&43.6&56.7&44.3\\
    SDNet + \textbf{Dual Attention}+\textbf{Knowledge Distillation}&\textbf{44.3}&\textbf{57.9}&\textbf{44.0}&\textbf{45.9}&\textbf{59.1}&\textbf{46.8}\\
    \midrule
    BERT-base \cite{devlin2018bert}&42.3&55.8&50.1&40.6&54.1&48.0\\
    BERT-base + sub-word unit \cite{li2018spoken}&43.2&56.8&51.1&41.6&55.4&48.9\\
    BERT-base+ SLU \cite{serdyuk2018towards}&42.5&56.1&50.3&41.0&54.6&48.1\\
    BERT-base + back-translation \cite{lee2018odsqa}& 42.9&56.5&50.5&41.5&55.2&48.6\\
    BERT-base + domain adaptation \cite{lee2019mitigating}& 43.1&57.0&51.0&41.7&55.7&49.0\\
    aeBERT \cite{kuo2020audio}&43.0&56.9&51.5&41.8&55.6&50.3\\
    BERT-base + \textbf{Dual Attention}&44.3&58.3&52.6&42.7&57.0&51.1\\
    BERT-base + \textbf{Knowledge Distillation}&44.1& 58.8&52.9& 42.8& 57.7&51.3\\
    BERT-base + \textbf{Dual Attention}+\textbf{Knowledge Distillation}&\textbf{46.5}& \textbf{61.1}&\textbf{55.1}& \textbf{45.6}& \textbf{60.1}&\textbf{53.6}\\
    \midrule
    ALBERT-base \cite{lan2019albert}&42.7&56.0&50.4&41.4&55.2&49.5\\
    ALBERT-base + sub-word unit \cite{li2018spoken}&43.7&57.2&51.2&42.6&56.8&50.3\\
    ALBERT-base + SLU \cite{serdyuk2018towards}&42.8&56.3&50.5&41.7&55.7&49.7\\
    ALBERT-base + back-translation \cite{lee2018odsqa}& 43.5&57.1&50.9&42.4&56.4&50.0\\
    ALBERT-base + domain adaptation \cite{lee2019mitigating}& 43.5&57.0&51.5&42.7&56.7&50.7\\
    ALBERT-base + \textbf{Dual Attention}&44.7&59.4&52.0&43.8&58.4&51.3\\
    ALBERT-base + \textbf{Knowledge  Distillation}&44.8&59.6&52.7&43.9&58.7&51.6\\
    ALBERT-base + \textbf{Dual Attention}+ \textbf{Knowledge  Distillation}&\textbf{47.3}&\textbf{61.9}&\textbf{55.5}&\textbf{46.1}&\textbf{61.3}&\textbf{53.6}\\
    \bottomrule
    \end{tabular}
    }
     \caption{Comparison of key components in \textsc{DDNet}. We denote the model trained on speech document and text corpus as the~\textit{teacher} model, and the one trained on the ASR transcripts as the~\textit{student} model.}
    \label{tab:my_label_3}
    \vspace{-15pt}
\end{table*}

\vspace{-0.5em}
\section{Experiments and Results}
\vspace{-0.5em}
In this section, we first describe the collection and filtering process of our proposed Spoken-CoQA dataset in detail. Next, we introduce several state-of-the-art language models as our baselines, and then evaluate the robustness of these models on our proposed Spoken-CoQA dataset. Finally, we provide a thorough analysis of different components of our method. Note that we use the default settings in all evaluated methods.

\myparagraph{Data Collection.}
\label{subsec:data}
We detail the procedures to build Spoken-CoQA as follows.  First, we select the conversational question-answering dataset CoQA \cite{reddy2019coqa}\footnote{Considering that the test set of CoQA \cite{reddy2019coqa} idoes not publicly availablesh the test set, we follow the widely used setting in the spoken question answering task \cite{li2018spoken}, where we divide Spoken-CoQA dataset into train and test set.} as our basis data since it is one of the largest public CQA datasets. CoQA contains around 8k stories (documents) and over 120k questions with answers. The average dialogue length of CoQA is about 15 turns, and the answers areis in free-form texts. In CoQA, the training set and the development set contain 7,199 and 500 conversations over the given stories, respectively. Therefore, we use the CoQA training set as our reference text of the training set and the CoQA development set as the test set in Spoken-CoQA.

Next, we employ the Google text-to-speech system to transform the questions and documents in CoQA into the spoken form, and adopt CMU Sphinx to transcribe the processed spoken contents into ASR transcriptions. In doing so, we collect more than 40G audio data, and the data duration is around 300 hours. The ASR transcription has a kappa score of 0.738 and Word Error Rates (WER) of 15.9$\%$, which can be considered sufficiently good since it is below the accuracy threshold of 30$\%$ WER \cite{gaur2016effects}. For the test set, we invite $5$ human native English speakers to read the sentences of the documents and questions. The sentences of one single document are assigned to a single speaker to keep consistency, while the questions in one example may have different speakers. All speech files are sampled at 16kHz, following the common approach in the speech community. We provide an example of our Spoken-CoQA dataset in Table \ref{tab:scqa_example} and Fig.~\ref{fig:mel}.

\myparagraph{Data Filtering}
\label{supp:datafilter}
In our SCQA task, the model predicts the start and end positions of answers in the ASR transcriptions. As a result, during data construction, it is necessary for us to perform data filtering by eliminating question-answer pairs if the answer spans to questions do not exist in the noisy ASR transcriptions.
We follow the conventional settings in \cite{lee2018odsqa}\footnote{We compare different Speech APIs, \eg, Google and CMU. Considering the quality of generated speech transcripts, we choose Google TTS for TTS and CMU Sphinx for ASR.}.
In our approach, an ASR question will be removed if the ground-truth answers do not exist in ASR passages. However, when coreference resolution and inference occurs, the contextual questions related to the previous ones are required to be discarded too. For the case of coreference resolution, we change the corresponding coreference. For the case of coreference inference, if the question has strong dependence on the previous one that has already been discarded, it will also be removed. 
After data filtering, we get a total number of our Spoken-CoQA dataset, we collect 4k conversations in the training set, and 380 conversations in the test set in our Spoken-CoQA dataset, respectively. Our dataset includes 5 domains, and we show the domain distributions in Table \ref{tab:stats}.

\myparagraph{Baselines.}
For SCQA tasks, our \textsc{DDNet} is able to utilize a variety of backbone networks for SCQA tasks. We choose several state-of-the-art language models (FlowQA \cite{huang2018flowqa}, SDNet \cite{zhu2018sdnet}, BERT-base \cite{devlin2018bert}, ALBERT \cite{lan2019albert}) as our backbone network baselines. We also compare our proposed \textsc{DDNet} with several state-of-the-art SQA methods \cite{lee2018odsqa,serdyuk2018towards,lee2019mitigating,kuo2020audio}. To use the \textit{teacher-student} architecture in our models, we first train baselines on the CoQA training set as \textit{teacher} and then evaluate the performances of testing baselines on CoQA dev set and Spoken-CoQA dev set. Finally, we train the baselines on the Spoken-CoQA training set as \textit{student} and evaluate the baselines on the CoQA dev set and Spoken-CoQA test set. We provide quantitative results in Table~\ref{tab:my_label_2}.

\myparagraph{Experiment Settings.}
We use the official BERT \cite{devlin2018bert} and ALBERT \cite{lan2019albert} as our textual embedding modules. We use BERT-base \cite{devlin2018bert} and ALBERT-base \cite{lan2019albert}, which both include 12 transformer encoders, and the hidden size of each word vector is 768. BERT and ALBERT both utilize BPE as the tokenizer, but FlowQA and SDNet use SpaCy \cite{spacy2} for tokenization. Under the circumstance when tokens in spaCy \cite{spacy2} correspond to more than one BPE sub-tokens, we average the BERT embeddings of these BPE sub-tokens as the final embeddings for each token. For fair comparisons, we use standard implementations and hyper-parameters of four baselines for training. The balancing factor $\alpha$ is set to 0.9, and the temperature~$\tau$ is set to 2. We train all models on 4 24GB RTX GPUs, with a batch size of 8 on each GPU. For evaluation, we use three metrics: Exact Match (EM), $F_1$ score and Audio Overlapping Score (AOS) \cite{li2018spoken}  to compare the model performance comprehensively. Please note that the metric numbers of baseline may be different from that in the CoQA leader board as we use our own implementations, Note that, we only utilize the {\em student} network for inference.

\begin{table}[h]
   
  
    \centering
    \footnotesize
    \resizebox{0.45\textwidth}{!}{%
    \begin{tabular}{l|cc|cc}
    \toprule
    & \multicolumn{2}{|c|}{\textbf{SQuAD dev}}&\multicolumn{2}{|c}{\textbf{S-SQuAD test}}\\
    \textbf{Methods}&EM&F1 &EM&F1\\
    \midrule\midrule
   FLowQA \cite{huang2018flowqa}&51.9&65.7&49.1&63.9\\
   FlowQA +\textbf{DA} &53.6&67.3&50.4&65.3\\
   FLowQA+ \textbf{KD} &53.5&67.3&50.9&65.8\\
   FLowQA+\textbf{DA}+ \textbf{KD} &\textbf{55.6}&\textbf{68.8}&\textbf{52.8}&\textbf{68.0}\\
    \hline
    SDNet \cite{zhu2018sdnet}&56.1&70.5&57.8&71.8\\
    SDNet + \textbf{DA}&58.3&71.4&59.3&73.8\\
    SDNet + \textbf{KD}&58.7&71.9&59.2&73.6\\
    SDNet + \textbf{DA}+ \textbf{KD}&\textbf{60.1}&\textbf{73.7}&\textbf{60.9}&\textbf{75.7}\\
    \hline
    BERT-base \cite{devlin2018bert}&58.3&70.2&58.6&71.1\\
    BERT-base + \textbf{DA}& 59.9 & 72.8 & 61.0 & 74.1\\
    BERT-base + \textbf{KD}& 60.1 & 72.2 & 60.8 & 73.8\\
    BERT-base + \textbf{DA}+ \textbf{KD}&\textbf{62.1}& \textbf{74.6}& \textbf{63.3}& \textbf{76.0}\\
    \hline
    ALBERT-base \cite{lan2019albert}&59.1&71.9&59.4&72.2\\
    ALBERT-base + \textbf{DA}&60.5&73.1&61.2&74.2   \\
    ALBERT-base + \textbf{KD}&60.8&73.6&61.9&74.7   \\
    ALBERT-base + \textbf{DA}+ \textbf{KD}&\textbf{62.6}&\textbf{75.7}&\textbf{64.1}&\textbf{77.1}\\
    \bottomrule
    \end{tabular}
    }
     \caption{Comparison of our method. We set the model on  text corpus as the~\textit{teacher} model, and the one on the ASR transcripts as the~\textit{student} model. DA and KD represent Dual Attention and knowledge distillation.}
       \label{tab:my_label_4}
        \vspace{-20pt}
\end{table}

\myparagraph{Results.}
We compare several \textit{teacher-student} pairs on CoQA and Spoken-CoQA dataset and the quantitative results are shown in Table~\ref{tab:my_label_2}. We can observe that the average F1 scores is 77.6$\%$ when training on CoQA (text document) and testing on CoQA dev set. However, when training the models on Spoken-CoQA (ASR transcriptions) and testing on Spoken-CoQA test set, the average F1 scores drops significantly to 49.3$\%$. For FlowQA, the performance even drops by 40.4 pts in terms of F1 score. This corroborates the importance of mitigating ASR errors.

Table~\ref{tab:my_label_3} compares our approach \textsc{DDNet} to all the previous results. As shown in the table, our distillation models achieve strong performance, and incorporating DA mechanism further improves the results considerably. Our \textsc{DDNet} using BERT-base models as backbone achieves similar or better results compared to all the state-of-the-art methods, and we observe that using a larger encoder ALBERT-base will give further bring large gains on performance.

As seen from Table \ref{tab:my_label_4}, we find that our best model ALBERT-base \textit{only} trained with KD achieve an absolute EM/F1 improvement of +1.7pts/+1.7pts, +2.5pts/+2.5pts, on CoQA and S-CoQA, respectively.
 This shows that cross-modal information is useful for the model, hence demonstrating that such information is able to build more robust contextualized cross-modal representations for the network performance improvements.
As shown in Table \ref{tab:my_label_4}, we also observe that our approach ALBERT-base \textit{only} trained with DA outperforms the original method by an absolute EM/F1 of +1.4pts/+1.2pts, +1.8pts/+2.0pts, on CoQA and S-CoQA, respectively.
This indicates that the fine-grained alignment between audio and text learned through DA during training benefits the downstream speech-and-language tasks.
Overall, our results suggest that such a network notably improves prediction performance for spoken conversational question answering tasks. Such significant improvements demonstrate the effectiveness of~\textsc{DDNet}.

\section{Ablation Study}
We conduct ablation studies to show the effectiveness of several components
in DDNet in this section and appendix.

\myparagraph{Multi-Modality Fusion Mechanism.}
To study the effect of different modality fusion mechanisms, we introduce a novel fusion mechanism \textit{Con Fusion}: first, we directly concatenate two output embedding from speech-BERT and text-BERT models, and then pass it to the encoding layer in the following SCQA module. In Table \ref{tab:my_label_5}, we observe that Dual Attention mechanism outperform four baselines with \textit{Con Fusion} in terms of EM and F1 scores. We further investigate the effect of uni-model input. Table \ref{tab:my_label_5} shows that~\textit{text-only} performs better than~\textit{speech-only}. One possible reason for this performance is that only using speech features can bring additional noise. Note that speech-only (text-only) means that we only feed the speech (text) embedding for speech-BERT (text-BERT) to the encoding layer in the SCQA module.

\section{Conclusions}
In this paper, we have presented SCQA, a new spoken conversational question answering task, for enabling human-machine communication. We make our effort to collect a challenging dataset - Spoken-CoQA, including multi-turn conversations and passages in both text and speech form. We show that the performance of existing state-of-the-art models significantly degrade on our collected dataset, hence demonstrating the necessity of exploiting cross-modal information in achieving strong results. We provide some initial solutions via knowledge distillation and the proposed dual attention mechanism, and have achieved some good results on Spoken-CoQA. Experimental results show that \textsc{DDNet} achieves substantial performance improvements in accuracy.
In future, we will further investigate the different mechanisms of integrating speech and text content, and our method also opens up the possibility for downstream spoken language tasks.

\bibliography{anthology,custom}
\bibliographystyle{acl_natbib}

\appendix

\section*{Appendix}
\label{sec:appendix}

\section{Temperature~$\tau$}
\begin{figure*}[t]
    \centering
    \includegraphics[width=0.95\textwidth]{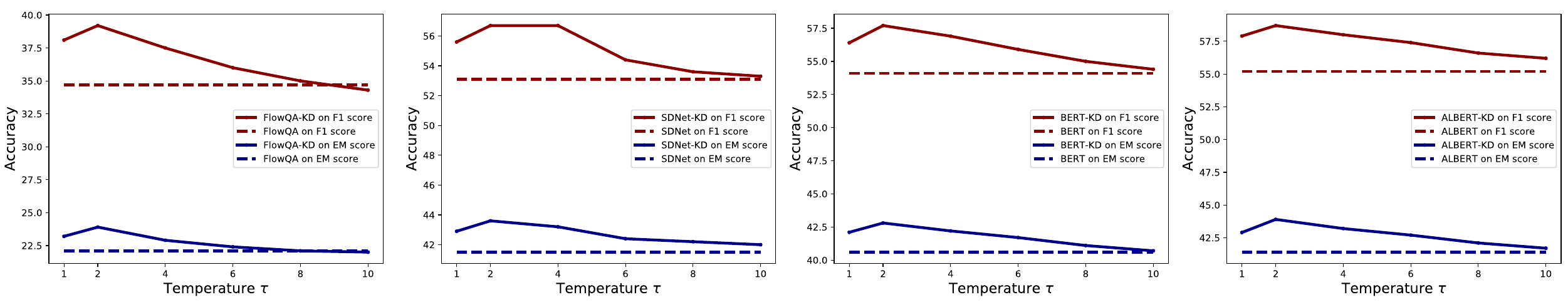}
    \vspace{-10pt}
    \caption{Ablation studies of temperature $\tau$ on \textsc{DDNet} performance~(FlowQA, SDNet, BERT, ALBERT). Red and blue denote the results on  Spoken-CoQA test set.}
    \label{fig:t}
    \vspace{-15pt}
\end{figure*}

To study the effect of temperature~$\tau$, we conduct the additional experiments of four baselines with the standard choice of the temperature~$\tau \in \{1,2,4,6,8,10\}$. All models are trained on Spoken-CoQA dataset, and validated on the Spoken-CoQA test set. We present the results in Figure~\ref{fig:t}. When $\tau$ is set to 2, four baselines all achieve their best performance in term of F1 and EM metrics.

\section{Effects of Different Word Error Rates}
\begin{figure}[h]
    \centering
    \includegraphics[width=0.4\textwidth]{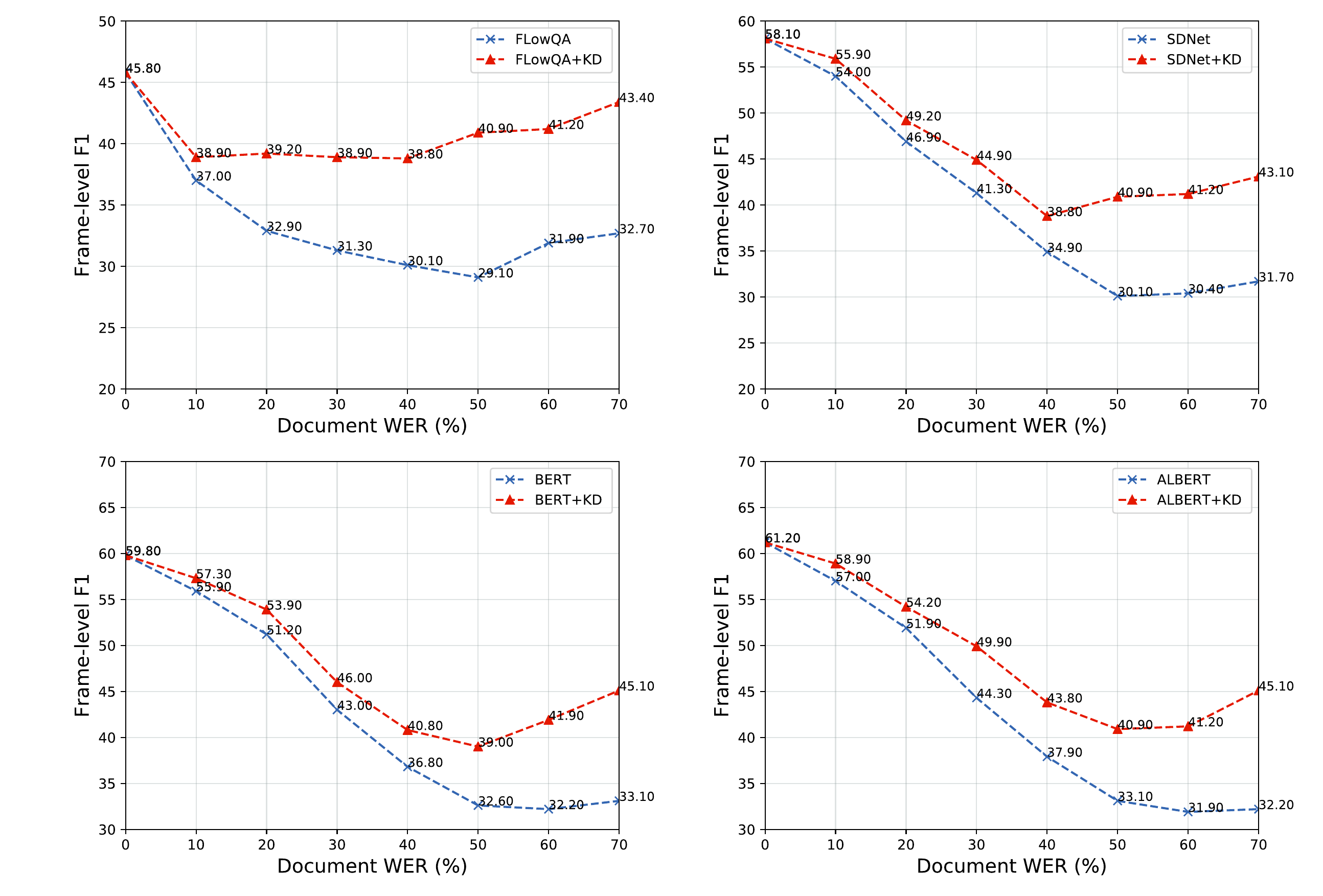}
    \vspace{-10pt}
    \caption{Comparison of different WER on Spoken-CoQA.}
    \vspace{-15pt}
    \label{fig:wer}
\end{figure}

We study how the network performances change when trained with different word error rates (WER) in Figure \ref{fig:wer}. Specifically, we first split Spoken-SQuAD and Spoken-CoQA into smaller groups with different WERs. Then we utilize Frame-level F1 score \cite{chuang2019speechbert} to validate the effectiveness of our proposed method on Spoken-CoQA. In Figure \ref{fig:wer}, we find that all evaluated networks for two tasks are remarkably similar: all evaluated models suffer larger degradation in performance at higher WER, and adopting knowledge distillation strategy is capable of alleviating such issues. Such phenomenon further demonstrates the importance of~\textit{knowledge distillation} in the case of high WER.

\section{Results on Human Recorded Speech}
The results using BERT-base as the baseline are shown in Table \ref{table:human}. We train the model in the Spoken-CoQA training dataset and evaluate the model in both machine synthesized and human recorded speech. As shown in Table \ref{table:human}, the average EM/F1/AOS scores using BERT fell from 40.6/54.1/48.0 to 39.4/53.0/46.8, respectively. In addition, the similar trends can be observed on our proposed method. We hypothesise that the human recorded speech introduces additional noise during training, which leads to the performance degradation.

\begin{table}[h]
    \caption{Comparisons between human recorded speech and synthesized speech. We employ BERT as our base model.}
    \vspace{-10pt}
    \label{table:human}
    \centering
    \resizebox{0.5\textwidth}{!}{%
    \begin{tabular}{c c c c c}
    \toprule
{Dataset} & Method & EM & F1 & AOS \\ \hline\hline
\multirow{2}{*}{Machine} & BERT & 40.6 & 54.1 & 48.0 \\
& BERT+KD+DA & 45.6 & 60.1 & 53.6 \\ \hline
\multirow{2}{*}{Human} & BERT & 39.4 & 53.0 & 46.8 \\
& BERT+KD+DA & 44.7 & 59.4 & 53.1 \\
\bottomrule
    \end{tabular}
    }
\end{table}

\begin{table}[t]
    \centering
\caption{Comparison of different fusion mechanisms in \textsc{DDNet}.}
\vspace{-5pt}
    \resizebox{0.48\textwidth}{!}{%
    \begin{tabular}{l|cc|cc}
    \toprule
    & \multicolumn{2}{c}{\textbf{CoQA dev}}&\multicolumn{2}{|c}{\textbf{S-CoQA test}}\\
    \textbf{Models}&EM&F1 &EM&F1\\
    \toprule\toprule
    FlowQA \cite{huang2018flowqa}&40.9&51.6&22.1&34.7\\
    \quad+ \textit{speech-only} &40.8&51.2&21.8&34.0\\
    \quad+ \textit{text-only} &41.1&51.7&22.4&35.3\\
    \quad+ \textit{Con Fusion} &41.0&52.0&22.1&35.2\\
    \quad+ \textbf{Dual Attention}&\textbf{42.3}&\textbf{53.0}&\textbf{23.5}&\textbf{38.8}\\
    \hline
    SDNet \cite{zhu2018sdnet}&40.1&52.5&41.5&53.1\\
    \quad+ \textit{speech-only}&39.3&51.6&40.9&52.28\\
    \quad+ \textit{text-only}&40.2&52.7&41.5&53.3\\
    \quad+ \textit{Con Fusion}&40.3&52.6&41.5&53.2\\
    \quad+ \textbf{Dual Attention}& \textbf{41.7}&\textbf{55.2}&\textbf{43.2}&\textbf{56.1}\\
    \hline
    
    BERT-base \cite{devlin2018bert}&42.3&55.8&40.6&54.1\\
    \quad+ \textit{speech-only}&41.9& 55.8& 40.2& 54.1\\
    \quad+ \textit{text-only}&42.4& 56.0& 40.9& 54.3\\
    \quad+ \textit{Con Fusion}&42.3& 56.0& 40.8& 54.1\\
    \quad+ \textbf{Dual Attention}&\textbf{44.3}&\textbf{58.3}&\textbf{42.7}&\textbf{57.0}\\
   
    \hline
    ALBERT-base \cite{lan2019albert}&42.7&56.0&41.4&55.2\\
   \quad+ \textit{speech-only}&41.8&55.9&41.1&54.8\\
    \quad+ \textit{text-only}&42.9&56.3&41.4&55.7\\
    \quad+ \textit{Con Fusion}&42.7&56.1&41.3&55.4\\
    \quad+ \textbf{Dual Attention}&\textbf{44.7}&\textbf{59.4}&\textbf{43.8}&\textbf{58.4}\\
    \bottomrule
    \end{tabular}
    }
\label{tab:my_label_5}
\vspace{-10pt}
\end{table}

\section{More Information about Spoken-CoQA}
\label{supp:morecompare-scoqa}
\begin{figure*}[h]
    \centering
    \includegraphics[width=0.95\textwidth]{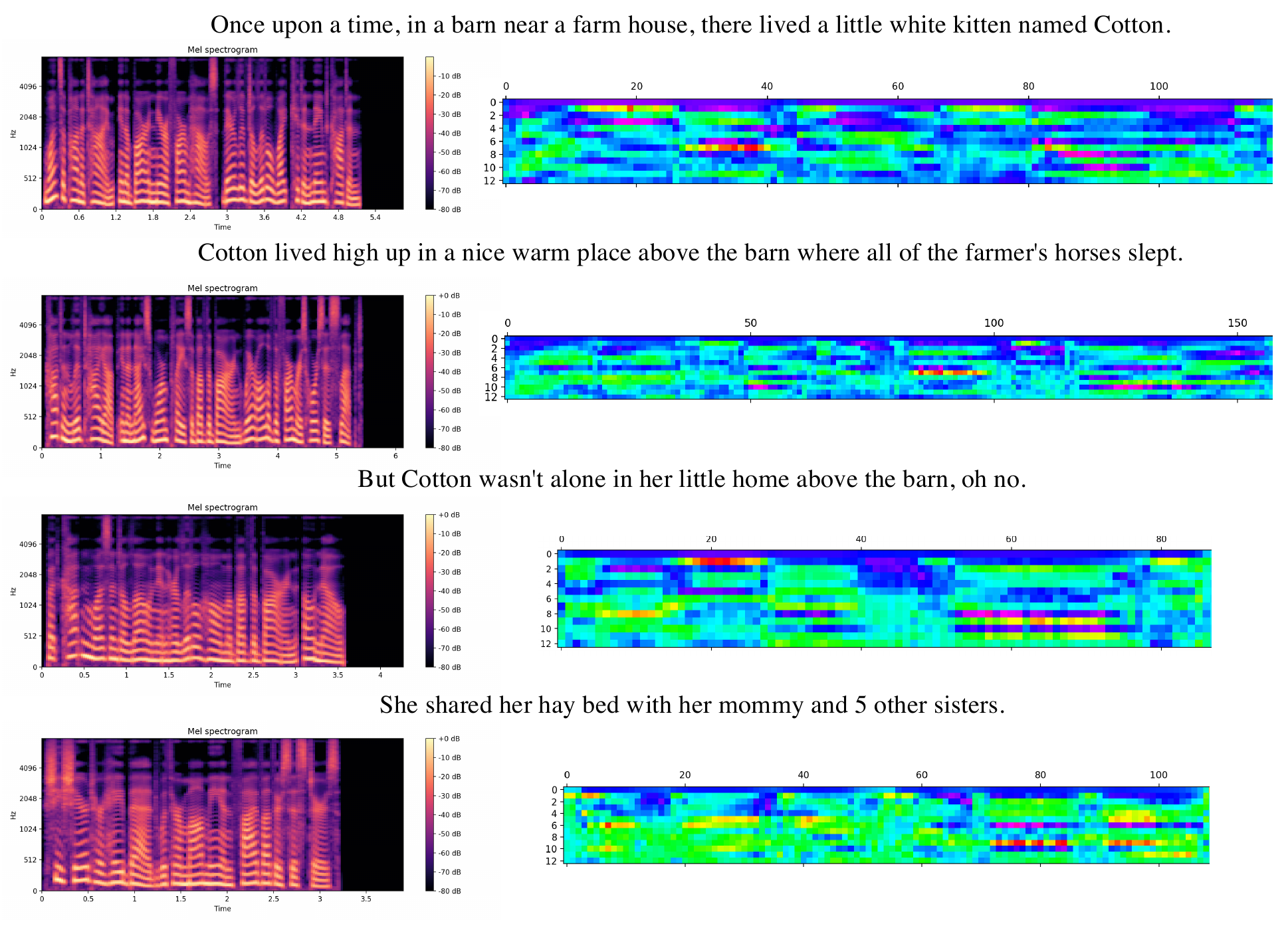}
    \caption{Examples of the log-mel spectrograms and the corresponding MFCC feature embedding. It can see that the log-mel spectrograms corresponds to different example sentences from the Spoken-CoQA dataset.}
    \label{fig:mel}
\end{figure*}

To perform qualitative analysis of speech features, we visualize the log-mel spectrogram features and the mel-frequency cepstral coefficients (MFCC) feature embedding learned by \textit{DDNet} in Figure~\ref{fig:mel}. We can observe how the spectrogram features respond to different sentence examples. In this example, we observe that given the text document~(ASR-document), the conversation starts with the question~$Q_1$~(ASR-$Q_1$), and then the system requires to answer $Q_1$ (ASR-$Q_1$) with $A_1$ based on a contiguous text span $R_1$. Compared to the existing benchmark datasets, ASR transcripts~(both the document and questions) are much more difficult for the machine to comprehend questions, reason among the passages, and even predict the correct answer.

\section{More Comparisons on Spoken-SQuAD}
\label{supp:morecompare-squad}
To verify that our proposed \textsc{DDNet} is not biased towards specific settings, we conduct a series of experiments on Spoken-SQuAD \cite{li2018spoken} by training the \textit{teacher} model on textual documents, and the \textit{student} model on the ASR transcripts. From the Table \ref{tab:my_label_3}, compared with the performances on Spoken-CoQA, all baselines performances improve by a large margin, indicating our proposed dataset is a more challenging task for current models. We verify that, in the setting (KD+DA), the model consistently achieves significant performance boosts on all baselines. Specifically, for FlowQA, our method achieves 55.6$\%$/68.8$\%$ (vs.51.9$\%$/65.7$\%$), and 52.8$\%$/68.0$\%$ (vs.49.1$\%$/63.9$\%$) in terms of EM/F1 score over the text documents and ASR transcriptions, respectively. For SDNet, our method outperforms the baseline without distillation, achieving 60.1$\%$/73.7$\%$ (vs.56.1$\%$/70.5$\%$) and 60.9$\%$/75.7$\%$ (vs.57.8$\%$/71.8$\%$) in terms of EM/F1 score. As for two BERT-based models (BRET-large and ALBERT-large), our methods with KD consistently improve EM/F1 scores to 62.1$\%$/74.6$\%$ (vs.58.3$\%$/70.2$\%$) and 63.3$\%$/76.0$\%$ (vs.58.6$\%$/71.1$\%$); 62.6$\%$/75.7$\%$ (vs.59.1$\%$/71.9$\%$) and 64.1$\%$/77.1$\%$ (vs.59.4$\%$/72.2$\%$), respectively. These results confirm the importance of knowledge distillation strategy and dual attention mechanism.

\section{Broader Impact}
In this section, we acknowledge that our work will not bring potential risks to society considering the data is from open source with no private or sensitive information. We also discuss some limitations of our work. First, we admit that using Google TTS for TTS and CMU Sphinx for ASR may affect the distribution of errors compared with the human recorded speech. Second, we currently cover only English language but it would be interesting to see that contributions for other languages would follow. Finally, as our collection comes with reliable data, it should trigger future analysis works on analyzing spoken conversational question answering biases.

\end{document}